# OSNet & MNetO: Two Types of General Reconstruction Architectures for Linear Computed Tomography in Multi-Scenarios

Zhisheng Wang, Zihan Deng, Fenglin Liu, Yixing Huang, Haijun Yu and Junning Cui

*Abstract*—Recently, linear computed tomography (LCT) systems have actively attracted attention as special imaging instruments different from conventionally rotated CT. To weaken projection truncation and image the region of interest (ROI) for LCT, the backprojection filtration (BPF) algorithm is an effective solution. However, in BPF for LCT, it is difficult to achieve stable interior reconstruction, and for differentiated backprojection (DBP) images of LCT, multiple rotation-finite inversion of Hilbert transform (Hilbert filtering)-inverse rotation operations will blur the image. To satisfy multiple reconstruction scenarios for LCT, including interior ROI, complete object, and exterior region beyond field-of-view (FOV), and avoid multiple rotation operations accompanying before and after Hilbert filtering, we propose two types of reconstruction architectures. The first overlays multiple DBP images to obtain a complete DBP image, then uses a network to learn the overlying or composite Hilbert filtering function, referred to as the Overlay-Single Network (OSNet). The second uses multiple networks to train different directional Hilbert filtering models for DBP images of multiple linear scanning trajectories, respectively, and then overlays the reconstructed results, i.e., Multiple Networks Overlaying (MNetO). In two architectures, we introduce a Swin Transformer (ST) block to the generator of pix2pixGAN to extract both local and global features from DBP images at the same time. We investigate two architectures from different networks, FOV sizes, pixel sizes, number of projections, geometric magnification, and processing time. Experimental results show that two architectures can both recover images. OSNet outperforms BPF in various scenarios. For the different networks, ST-pix2pixGAN is superior to pix2pixGAN and CycleGAN. MNetO exhibits a few artifacts due to the differences among the multiple models, but any one of its models is suitable for imaging the exterior edge in a certain direction.

This work was supported in part by the National Natural Science Foundation of China (Grant No.: 52075133), CGN-HIT Advanced Nuclear and New Energy Research Institute (Grant No.: CGN-HIT202215). (*Corresponding authors: Haijun Yu, Junning Cui. These authors contributed equally: Zhisheng Wang, Zihan Deng*).

Z. Wang and Z. Deng are with the Center of Ultra-precision Optoelectronic Instrument Engineering, Harbin Institute of Technology, Harbin, 150080, China.

F. Lin is with the Key Laboratory of Optoelectronic Technology and Systems, Ministry of Education, Chongqing University, Chongqing 400044, China.

Y. Huang is with Department of Radiation Oncology, University Hospital Erlangen, Friedrich-Alexander-University Erlangen-Nuremberg, 91054 Erlangen, Germany.

H. Yu is with the Key Laboratory of Optoelectronic Technology and Systems, Ministry of Education, Chongqing University, Chongqing 400044, China (e-mail: yuh1@cqu.edu.cn).

J. Cui is with the Center of Ultra-precision Optoelectronic Instrument Engineering, Harbin Institute of Technology, Harbin, 150080, China (e-mail: cuijunning@hit.edu.cn).

## I. INTRODUCTION

Recently, as special imaging instruments different from conventional rotated CT, linear computed tomography (LCT) systems have actively attracted a great deal of attention [1]–[14]. This is because rotated CT scanning modes are usually not the most suitable for some special scenarios and requirements, such as when the scanned objects change radically and require at least one of the following: high image quality, low cost, high scanned speed, enlarged field of view (FOV), security inspection, in service inspection, etc. For any one of the above-mentioned special scenes, a certain LCT system can indeed meet the requirement, such as: parallel translational CT (i.e., PTCT, ultra-low-cost medical imaging) [3], symmetric-geometry CT (SGCT, ultra-fast scanning) [4], and source translation CT (STCT, extending the FOV) [5]. In fact, these LCT systems all share a common characteristic: the linear scanning trajectory.

LCT often faces projection truncation issues due to the requirement of regions of interest (ROI) or high-resolution imaging, and backprojection filtration (BPF) is usually a preferred and effective reconstruction algorithm to solve this problem. In the differential backprojection (DBP) of BPF for LCT, the reconstructed DBP image is incompletely distributed along a single linear trajectory, i.e., with limited-angle artifacts. Therefore, multiple linear trajectories with different view angles are usually introduced to acquire the complete projection data within the FOV of the LCT. To further recover the object space, the finite inversion of Hilbert transform (here simply referred to as Hilbert filtering) for the DBP image of LCT needs to be performed along a cluster of lines parallel to a linear trajectory. In this way, it requires multiple different directional Hilbert filtering for DBP images to complete the LCT reconstruction task. In practice, because Hilbert filtering of the image matrix can only be performed in the orthogonal direction (e.g. the row or column direction of the matrix), it is necessary to perform forward and reverse rotation of the DBP image before and after this operation [6], [14]. However, in the reconstruction with multiple linear scanning trajectories, multiple rotation operations accompanying before and after Hilbert filtering inevitably cause the final image to lose some high-frequency details, whereas the higher spatial resolution has always been the unremitting and unified pursuit of all CT systems. Although BPF is somewhat imperfect here, it enables an exact reconstruction for commonly requested interior tomography if the value of any ROI sub-region is known [15], [16]. Unfortunately, this given sub-region is difficult to find in practical applications.

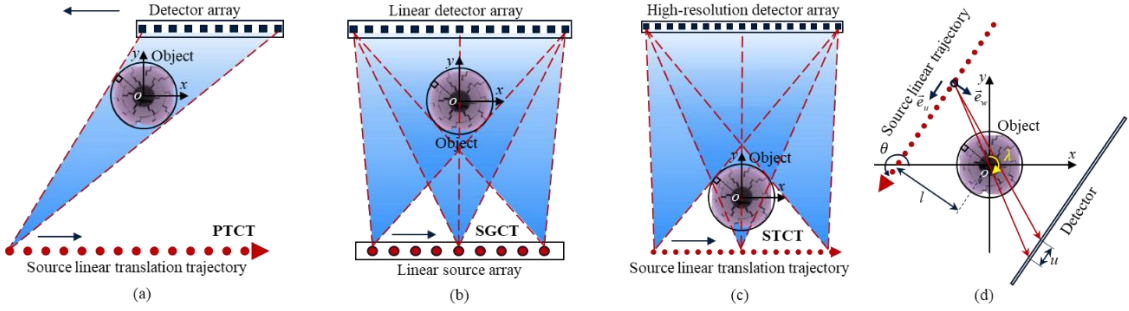

**Fig. 1.** Illustration of some different LCT imaging models, including (a) PTCT, (b) SGCT, (c) STCT, and (d) the general LCT geometry. For PTCT, the source and detector scan in a relatively parallel direction. For SGCT, many source foci are equidistantly distributed along a straight line and are sequentially fired. STCT is designed for micro-CT and is compatible with extending FOV and high-resolution imaging. For STCT, the object is placed close to the source to obtain large magnification, whereas the object is far away from the source in SGCT.

Compared to BPF for LCT, the filtered backprojection (FBP) type algorithm can avoid such rotated filtering operation as it directly processes the projection in the row. However, its global ramp-filter operator cannot directly handle the projection truncation, i.e., producing serious truncation artifacts at the truncated point and propagating errors to the entire image. Based on the FBP algorithm, sinogram extrapolation is a simple but approximate method to reduce artifacts by extrapolating the truncated data via a smooth function [17]. Iterative algorithms like the total variation (TV) penalty can uniquely reconstruct an interior ROI image when the function space of the measured object is composed of piecewise constant Sor smooth polynomial areas [18], [19]. Moreover, analytical and iterative algorithms can also be combined to accurately recover the high-frequency and low-frequency signals in ROI [11]. In fact, the computational complexity of iterative algorithms is still the main reason hindering their practical applications [21]. The principle of extra scanning is based on the fact that the exterior projections outside the ROI still generate non-zero local data inside the ROI. By removing the contribution of exterior regions from the truncated projection, an artifact-free ROI image can theoretically be obtained, but this will reduce scanning efficiency and introduce some errors [22]–[25].

Recently, deep learning networks have shown great potential in various applications, and various network architectures have achieved significant success in CT image reconstruction applications [21], [26]–[33]. These deep learning applications have been proven to gain good performance on noise suppression and structural fidelity and surpass previous iterative methods in reconstruction time [21], [32]. In theory, deep learning is demonstrated to be useful for inverse problems because it is closely related to a novel combinatorial signal representation using nonlocal basis convolved with data-driven local basis [21], [34], [35]. Previously, based on a modified architecture of the U-Net network, Han et al. [21] indicated that this network can learn the null space of FBP-reconstructed images with ring artifacts, but it lacks generalization, and some artifacts are displayed in the results. Meanwhile, this network has demonstrated great performance for DBP images in interior tomography with arbitrary ROI sizes. Unlike the only single-directional filtering in a circular scanning DBP image in Han's work, the direction needed to complete the Hilbert filtering of each DBP image is parallel to its linear trajectory in LCT, i.e., multiple different directional Hilbert filtering operations are requested [14].

In this paper, we combine the DBP formula of LCT with deep learning technology to implement directional finite inversion of Hilbert transform, i.e., without any rotation-Hilbert filtering-inverse rotation operation, aiming to satisfy high-quality multi-scenarios for LCT, including interior ROI, exterior edge, and complete object. We mainly make contributions as follows:

1) We present two types of general reconstruction network architectures for LCT. Theoretically, they provide solutions for almost all special LCT systems in multi-scenario' reconstruction. The primary distinction between these two architectures lies in the order of DBP image overlay and network model training. For convenience, the first architecture is referred to as Overlay-Single Network (OSNet). Especially, first it overlays DBP images from multiple linear scans to obtain a complete DBP image without limited-angle artifacts, then uses a network to learn the Hilbert filtering of superposition. The second architecture is using multiple networks to train different directional inversion of Hilbert transform models for DBP images of multiple linear scanning, respectively, and then overlaying multiple reconstructed results, i.e., Multiple Networks Overlaying (MNetO).

2) In the architectures mentioned above, we introduce a Swin Transformer (ST) block to the generator of the traditional pix2pixGAN network, which helps extract both local and global features from DBP images of LCT at the same time by using multiple sliding windows and self-attention coding. This new method successfully solves problems like the "global" pixel misalignment in DBP images of the LCT scanning mode. It also improves upon the traditional U-net networks by addressing issues related to the loss of "translation invariance" and enhancing the quality of image reconstruction.

3) Regarding experimental outcomes, our proposed OSNet architecture, where the Swin Transformer block is introduced into pix2pixGAN, outperforms BPF and traditional pix2pixGAN under various reconstruction scenes, including FOV sizes, pixel sizes, projection numbers, and geometric magnifications, when it comes to achieving superior reconstruction effects. MNetO surpasses BPF in terms of resolution during single-segment linear scanning reconstruction, so it is more widely adopted to complete high-quality exterior-edged reconstruction. Besides, both architectures demonstrate enhanced efficiency for large-size image reconstructions.

The rest of this paper is organized as follows. In Section 2, the LCT imaging geometry and the corresponding BPF type algorithm are reviewed. Then, Section 3 describes the reconstruction methods, which are followed by experimental results in Section 4. Finally, Sections 5 and 6 give the discussion and conclusion of this work, respectively.

## II. THEORY

### A. Linear computed tomography geometry

Figs. 1(a)–(c) describe the imaging geometries of some typical LCT systems. A global coordinate system $o$-$xy$ is attached to the imaging object. Without loss of generality, referring to the literature [4], [6]–[13], such scanning trajectories can be briefly described as general linear geometry [2], as shown in Fig. 1(d).

$$\vec{S}(\theta,\lambda) = \left(\frac{-l}{\cos(\theta+\lambda)}\sin\lambda, \frac{-l}{\cos(\theta+\lambda)}\cos\lambda\right), \quad (1)$$

where $\lambda$ is an angle from the positive $y$-axis to the vector from the source to origin, $\lambda \in (-\pi/2-\theta, \pi/2-\theta)$, $l$ is the source-to-object distance, and $\theta$ is an angle from the positive $x$-axis to the linear trajectory.

### B. BPF reconstruction for LCT

In the first step of the BPF algorithm for LCT, the general formula yielding an intermediate DBP image can be written as

$$b_\eta^\theta(\vec{r}) = \int_{\lambda_s}^{\lambda_e} \frac{1}{|\vec{r}-\vec{S}(\theta,\lambda)|} \cdot \frac{\partial p_\theta(\lambda,u)}{\partial u}\bigg|_{u=u^*} d\lambda, \quad (2)$$

where $\lambda_s$ and $\lambda_e$ are the angles of the starting and ending points of the source on the linear trajectory, respectively. $u^*$ is the $u$-coordinate of the ray passing through the reconstructed point $\vec{r} := (x,y)$ and reaching the detector. $p_\theta(\lambda,u)$ is the projection data of the current linear trajectory.

To translate from a DBP image $b_\eta^\theta(x,y)$ to an objective image $f_\theta(x,y)$, it is necessary to run a finite inversion of Hilbert transform $\mathcal{H}_\eta^{-1}$, i.e. $f_\theta(x,y) = -\mathcal{H}_\eta^{-1}(b_\eta^\theta(x,y)/2\pi)$ [36]. Theoretically, the finite inversion of Hilbert transform for a DBP image needs to follow PI-lines or chords, whereas the concept of PI-lines needs to be modified in LCT, as all PI-lines are on the same straight line parallel to the linear trajectory [2], [6], [37]. Here, within the finite range $r$ and all $t \in [L_r, U_r]$

(where $r = x\cos\eta + y\sin\eta$ , and $t = -x\sin\eta + y\cos\eta$ ), if $\mathcal{H}_\eta f(r\vec{\eta}+t\vec{\eta}^\perp)$ is known ( $\mathcal{H}_\eta f(x,y)$ denotes the Hilbert transform of $f(x,y)$ along a cluster of lines parallel to the $t$-axis) [14], where the angle between the positive $t$-axis and the $y$-axis is $\eta$, as shown in Fig. 2), $\mathcal{H}_\eta^{-1}$ is written as

$$\mathcal{H}_\eta^{-1}\left(\mathcal{H}_\eta f(r\vec{\eta}+t\vec{\eta}^\perp)\right) = f(r\vec{\eta}+t\vec{\eta}^\perp) = \frac{-1}{\sqrt{(t-L_r)(U_r-t)}}$$
$$\cdot \left(\int_{L_r}^{U_r} \sqrt{(t'-L_r)(U_r-t')}\frac{\mathcal{H}_\eta f(r\vec{\eta}+t'\vec{\eta}^\perp)}{\pi(t-t')}dt' + \mathcal{C}_r\right), (3)$$

where $\mathcal{C}_r$ is the offset function that needs to be analytically calculated. One simple method is to use a region that is known to be zero to calculate $\mathcal{C}_r$ to obtain a unique solution, whereas this condition is difficult to satisfy for interior reconstruction [36]. In addition, DBP images of LCT have different directions parallel to the linear trajectories, finite inversion of Hilbert transform with an arbitrary angle can be implemented, as shown in Fig. 2.

$$\mathcal{H}_\eta^{-1}(f_\theta(x,y)) = \mathcal{R}_\eta \mathcal{H}_0^{-1} \mathcal{R}_{-\eta}(f_\theta(x,y)), \quad (4)$$

where $\mathcal{R}_\eta$ is the rotation transform, $\eta = \theta - \pi/2$ [14].

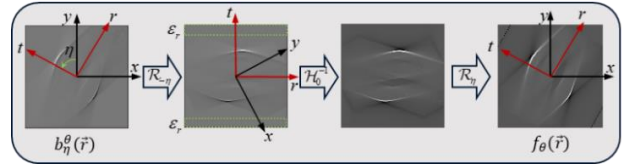

**Fig. 2.** Illustration of the finite inversion of Hilbert transform for one-segment arbitrary angular linear scanning in LCT.

$$f(x,y) = \sum_{i=1}^T f_{\theta_i}(x,y), \quad (5)$$

where we translate the variables $\theta$ to $\theta_i$, $\eta$ to $\eta_i$, and $f_{\theta_i}(x,y)$ represents the incomplete reconstructed result of the $i$-th linear scanning trajectory. In (4), the rotation-Hilbert filtering-inverse rotation operation is required for each-segment linear scanning, which inevitably reduces the spatial resolution of the final reconstructed image.

## III. METHOD

### A. Reconstruction architectures

By iteratively calculating the loss of pixels corresponding to the original image and label data, neural networks can learn the image transformation. Theoretically, any network with a sufficient scale can generate images that infinitely approximate the truth [38]. However, there are still problems when applying the usual loss functions for training. For instance, the widely used $\ell_2$ loss can lead to blurred prediction images. Therefore, a direct and effective solution is to choose a network model that can adaptively adjust the loss function to the diverse types of CT image data. The advanced pix2pixGAN network possesses these characteristics [39]. Besides, compared to unpaired networks and unsupervised learning, such as CycleGAN and Diffusion [40], pix2pixGAN with supervised learning under pairing conditions can generate "registered" images without significant pixel shift.

However, traditional pix2pixGAN typically employs a convolutional neural network (CNN)-based backbone network (e.g. Resnet [41], U-Net [42]) as the generator. The shallow structure of the network can only extract local feature information, and increasing its depth is necessary to obtain global features. Nevertheless, as the network becomes deeper, CNN will gradually lose "translation invariance" [43], which ultimately results in blurring or artifacts in the generated image. Considering the unique "global" "pixel drift" present in DBP images of LCT (Figs. 3(a)-(b)), we introduce a Swin Transformer block to the generator of the traditional pix2pixGAN network, i.e., referred to as ST-pix2pixGAN. With the conventional convolution generator replaced by the self-attention mechanism, we construct this conditional adductive generation network, ST-pix2pixGAN, that can address both global and local image feature extraction through location coding simultaneously. This approach aims to solve the distinctive problem of "long distance dependence" encountered during DBP image recovery.

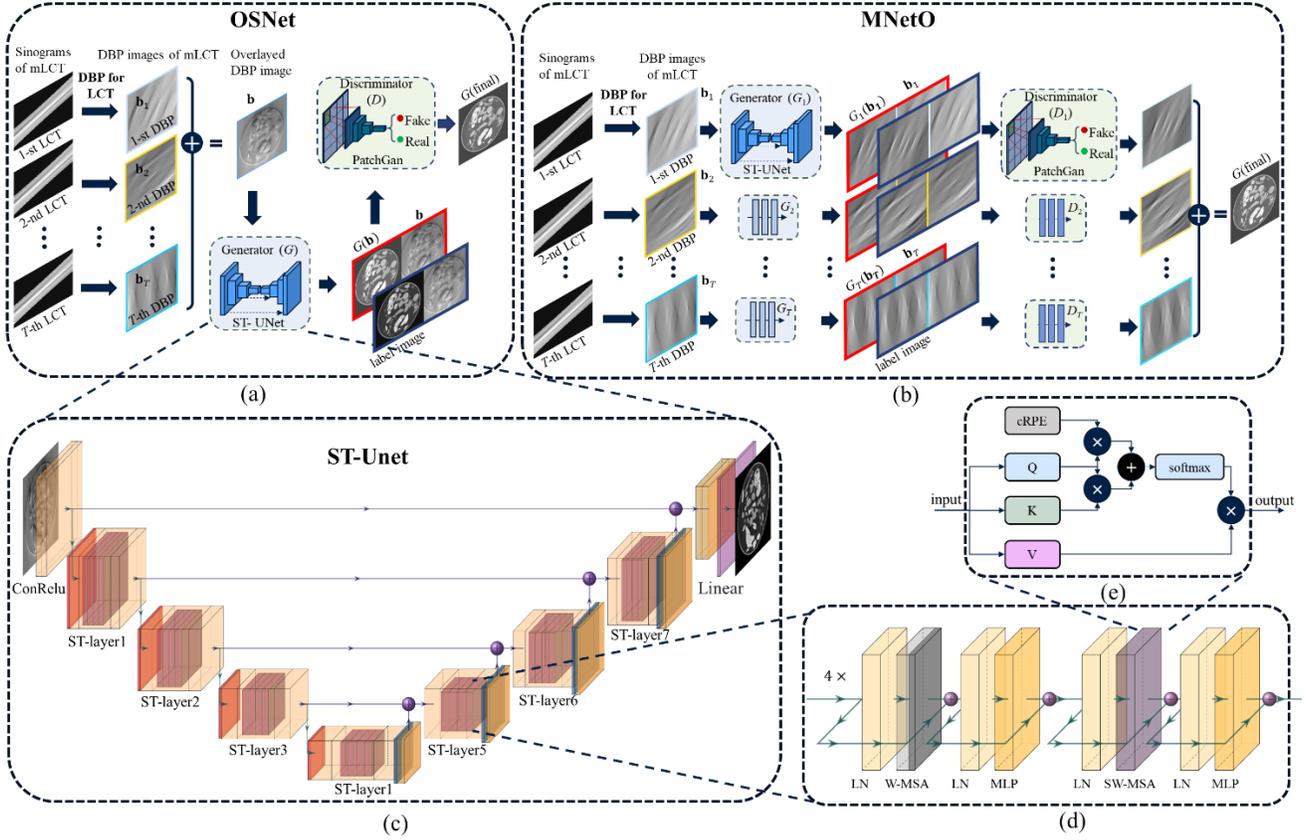

**Fig. 3.** Two types of reconstruction architectures for LCT. (a) OSNet reconstruction architecture; (b) MNetO reconstruction architecture; (c) ST-Unet structure in the generator in the proposed ST-pix2pixGAN network; (d) The structure of each ST-layer in Swin transformer Unet (ST-Unet); (e) The process of calculating the self-attention in each patch.

In LCT reconstruction, to translate multiple DBP images into the final complete reconstructed image without any rotation-Hilbert filtering-inverse rotation operation, we propose two types of deep learning-based general reconstruction architectures, including OSNet and MNetO. Fig. 3 illustrates how the ST-pix2pixGAN network is applied to our proposed OSNet and MNetO network architectures.

In OSNet (Fig.3(a)), we adopt a pix2pixGAN network $\mathcal{S}$ such that

$$\mathcal{S}\mathbf{b} = \mathbf{f}, \qquad (6)$$

where $\mathbf{b}$ denotes the overlapped DBP image of LCT, i.e. $\mathbf{b} = \sum_{i=1}^{T} \mathbf{b}_i$, and $\mathbf{b}_i$ is the DBP image of the $i$-th linear trajectory. $\mathbf{f}$ is the ground-truth image.

For reconstructing the ROI, the roles of the network $\mathcal{S}$ are: a) estimating the ROI size (and its restriction $[L_r + \varepsilon_r, U_r - \varepsilon_r]$, $\varepsilon_r$ is a small positive); b) forming the truncated DBP input to calculate the weighting; c) learning the kernel function for Hilbert transform and the offset $C_r$ [21]. This single network objective function can be described as follows:

$$\min_{\mathcal{S}} \sum_{j=1}^{M} \left\| \mathbf{f}^{(j)} - \mathcal{S}\mathbf{b}^{(j)} \right\|^2, \qquad (7)$$

where $\left\{ \mathbf{f}^{(j)}, \mathbf{b}^{(j)} \right\} \big|_{j=1}^{M}$ denotes that the training dataset consists of $M$-pairs (each pair includes ground-truth image and its overlapped DBP image of LCT), and ($j$) represents the $j$-th training data pair. To learn the Hilbert filtering for both complete DBP images and interior truncated DBP images, they as well as their corresponding ROI ground-truth images should be used as input and label data, respectively, for this training [21].

In MnetO (Fig. 3(b)) including $T$-segment models, after the DBP image $\mathbf{b}_i$ obtained for the $i$-th linear trajectory ($i = 1, 2, \ldots, T$), the $i$-th ST-pix2pixGAN network $\mathcal{S}_i$ is used to learn the Hilbert filtering with the angle $\eta_i$,

$$\mathcal{S}_i \mathbf{b}_i = \boldsymbol{\varphi}_i, \qquad (8)$$

where $\boldsymbol{\varphi}_i$ is the incomplete image with limited-angle artifacts of the $i$-th linear trajectory via the FBP-type algorithm under a large amount of source sample points without truncated projections, as the ground-truth image. The objective function for the $i$-th network is expressed as follows:

$$\min_{\mathcal{S}_i} \sum_{k=1}^{K} \left\| \boldsymbol{\varphi}_i^{(k)} - \mathcal{S}_i \mathbf{b}_i^{(k)} \right\|^2, \quad (9)$$

where $\left\{ \boldsymbol{\varphi}_i^{(k)}, \mathbf{b}_i^{(k)} \right\}_{i=1}^{K}$ denotes the training dataset consists of $K$-pairs (each pair includes the ground-truth image and its DBP image of the $i$-th linear trajectory, where $K = M * T$). Then, we perform superposition to obtain the final complete reconstructed image $\boldsymbol{\varphi}'$ within the FOV,

$$\boldsymbol{\varphi}' = \sum_{i=1}^{T} \boldsymbol{\varphi}_i', \quad (10)$$

and $\boldsymbol{\varphi}_i'$ denotes the output of the $i$-th trained model.

In Fig. 3(c), the generator $G$ in the ST-pix2pixGAN network replaces the convolution module in the traditional U-Net up-down sampling with the Swin Transformer layer (ST-layer) containing four Swin Transformor blocks. Patch merging method is used for downsampling and bilinear interpolation is used for upsampling, which transforms a DBP image into an image that approximates the label data. The process of up-down sampling can be described as:

$$\begin{cases} \mathcal{M}_{down_i} = L_{ST_i}(L_{PM_i} \mathcal{M}_{down_{i-1}})) \\ \mathcal{M}_{up_j} = L_{ST_j}(L_{B_j}(\mathcal{M}_{up_{j-1}} \oplus \mathcal{M}_{down_{N-j+1}})) \end{cases} \quad (11)$$

where $\mathcal{M}_{up_j}$ and $\mathcal{M}_{down_i}$ represent up and down sampling respectively, $L_{ST}$ represents Swin Transformer layer, $L_{PM_i}$ and $L_{B_j}$ represent up and down sampling and bottom layer respectively, and $\oplus$ represents channel stitching between images of the same size in the process of up and down sampling [44].

The structure of each ST-layer is shown in Fig. 3(d), which consists of the layer norm (LN), multilayer perceptron (MLP) and window multi-head self-attention module(W-MSA) and shifted window multi-head self-attention module (SW-MSA) [45]. In W-MSA and SW-MSA, set the patch size to 8*8, and calculate the self-attention in each patch according to the process shown in Fig. 3(e). The formula is as follows [46]:

$$\text{Attention} = \text{softmax}\left( \frac{Q \cdot K^T}{\sqrt{d_z}} + cPRE \right) V, \quad (12)$$

where $\sqrt{d_z}$ represents the scale factor when $Q$ and $K$ are scaled by dot product, $cPRE$ represents the relative position coding between different patches [47]. $V$ is the vector that makes the input value of the linear transformation.

The discriminator $D$ adopts PatchGAN. For OSNet, the total loss $\mathcal{L}_{G-OSNet}$ of $G$ is calculated using adversarial loss ($\mathcal{L}_{adv}$) and $\ell_1$ loss ($\mathcal{L}_{\ell_1}$) to measure the difference between the generated image $G(\mathbf{b})$ and the ground-truth image $\mathbf{f}$, as follows:

$$\mathcal{L}_{G-OSNet} = \mathcal{L}_{adv} + \mathcal{L}_{\ell_1} = -\lambda * \mathbb{E}_{\mathbf{b}}\big[log\big(D\big(G(\mathbf{b}), \mathbf{b}\big)\big)\big]$$
$$+ (1 - \lambda) * \mathbb{E}_{\mathbf{b}, \mathbf{f}}\big[\|\mathbf{f} - G(\mathbf{b})\|\big]. \quad (13)$$

Here, $\lambda$ is a coefficient ($\lambda \in [0, 1]$), which is used to balance the impact of the two losses on the generated image quality. $\mathcal{L}_{adv}$ is used to encourage $G$ to generate more realistic images, while

$\mathcal{L}_{\ell_1}$ is used to maintain the structural similarity between the generated images and the real images. $D$ loss uses binary cross-entropy loss to measure the difference between the classification results of $D$ and labels for the generated and real images. The loss of $D$ is the sum of the loss $\mathcal{L}_r$ for $\mathbf{f}$ with label 1 and the loss $\mathcal{L}_f$ for $G(\mathbf{b})$ with label 0, as follows:

$$\mathcal{L}_{D-OSNet} = \mathcal{L}_r + \mathcal{L}_f =$$
$$-\mathbb{E}_{\mathbf{b}, \mathbf{f}}\big[log\big(D(\mathbf{b}, \mathbf{f})\big)\big] - \mathbb{E}_{\mathbf{b}}\big[log\big(1 - D(\mathbf{b}, G(\mathbf{b}))\big)\big]. \quad (14)$$

For MNetO, since multiple models are used to learn the transformation space in different directions, its loss function is also more complex, For the generator loss $\mathcal{L}_{G-MNetO}$, it can be expressed in the following form:

$$\mathcal{L}_{G-MNetO} = \sum_{i=1}^{T}(\mathcal{L}_{adv} + \mathcal{L}_{\ell_1}) = -\lambda \sum_{i=1}^{T} \mathbb{E}_{\mathbf{b}}\big[log\big(D(G(\mathbf{b}), \mathbf{b})\big)\big]$$
$$+ (1 - \lambda) \sum_{i=1}^{T} \mathbb{E}_{\mathbf{b}}\big[\|\mathbf{f} - G(\mathbf{b})\|\big]. \quad (15)$$

Its discriminator loss can be expressed as:

$$\mathcal{L}_{D-MNetO} = \sum_{i=1}^{T}(\mathcal{L}_r + \mathcal{L}_f) = -\sum_{i=1}^{T} \mathbb{E}_{\mathbf{b}, \mathbf{f}}\big[log\big(D(\mathbf{b}, \mathbf{f})\big)\big]$$
$$- \sum_{i=1}^{T} \mathbb{E}_{\mathbf{b}}\big[log\big(1 - D(\mathbf{b}, G(\mathbf{b}))\big)\big]. \quad (16)$$

Therefore, the total loss function of ST-pix2pixGAN can be expressed as:

$$\mathcal{L}(G, D) = \min(\mathcal{L}_G) + \min(\mathcal{L}_D). \quad (17)$$

During the training, by alternating the training of $G$ and $D$, ST-pix2pixGAN can continuously optimize the quality of the generated images. To further investigate suitable networks, we also introduce CycleGAN [48] as comparisons.

### B. Datasets

We conducted experiments on the **2DeteCT** dataset [49] for our two networks and corresponding methods. The dataset is a large 2D scalable, trainable experimental micro-computed tomography dataset for machine learning recently released in 2023 by the Computational Imaging Group Information Science Center, XG 1098, Amsterdam, the Netherlands. It is suitable for our selected example: micro-CT-oriented STCT in the LCT systems. The dataset was obtained using a highly flexible, programmable and customized X-ray CT scanner developed by Texan-XRE, which collected data under the projection of a 90 kilovolt, 90 watt conebeam micro-focus X-ray source. The 2DeteCT dataset scanned a total of 5000 slices by reading the detector central row to simulate the geometry of the fan-shaped beam and looking for dried fruits and nuts filled with a cylindrical tube of similar density. In addition, the additional 750 out-of-distributed (OOD) slices from the sample, whose fillings, including screws that could produce metal artifacts, as well as other materials of different densities, were different from the above 5000 slices, had been collected to help users verify the generalization and robustness [49]. We took 5,000 raw images and divided them into a training set, a test set, and a verification set in a ratio of 8:1:1. The training set is also enhanced using data enhancement methods, including rotation, mirroring, and flipping operations, for training network parameters, while the validation set is used to determine the best hyperparameter Settings and prevent overfitting. Finally, 750 OOD images were used to test and evaluate the performance and generalization ability of the network.

In OSNet, we use the reconstructed image in the dataset as the reference label data. The DBP operator for LCT is applied to obtain multiple DBP images and superimpose them to form the complete DBP image, which served as the input for training the model. The original input for MNetO is the DBP image of each LCT, and the corresponding image obtained via V-FBP (whose source sample points per linear scanning are set to 2001) served as the label data.

In addition, we verify the robustness and generalization of our trained model of OSNet under different conditions, including different pixel sizes (here, we only set the number of pixels to be reconstructed 256, 512 and 1024), FOV sizes (we only need to change the length of linear trajectory $LS$ from 12 mm to 20 mm to implement according to Ref. [7]), number of projections (we set different source sampling points of each LCT, including 251, 1001, and 2001), and geometric magnification factors (i.e., changing the detector-to-source distances from 110 mm to 210 mm to indirectly adjust).

*C. Network training*

We implement OSNet and MNetO based on the Pytorch framework. We perform the model training, validation, and testing on the Intel Core i77700 (3.60 GHz) central processing unit (CPU) and eight RTX3090 graphics processing units (GPUs).

We use similar training parameters for the two types of networks, using Adam optimizer as the network optimizer for training, with an initial learning rate of 0.0002, gradually decreasing to 0 by exponential decay, and a total of 200 training rounds. The batch size is set to 16, and the Adam optimizer is used, with beta1 = 0.5 and beta2 = 0.999. Set the embedding dim to 96. Note that, because the original data trained in MNetO is a single DBP image, to minimize the loss of image information during the rotation stacking process, we padded the original image with edges at a padding rate of 0.5, resulting in an input size of $1536 \times 1536$ for MNetO.

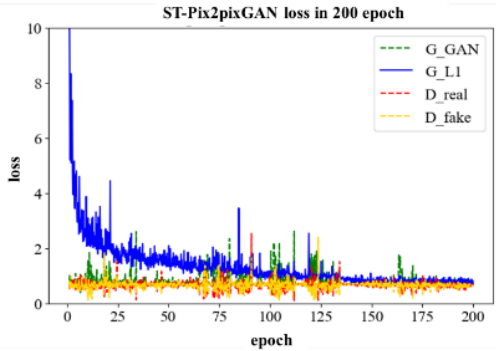

**Fig. 4.** Loss of network validation process.

Fig. 4 shows various losses during the training and validation of OSNet using ST-pix2pixGAN as an example. The objective function loss graph during the model validation phase. Since the $\ell_1$ loss curve eventually converges, we conclude that the network has been trained well and there is no overfitting.

## IV. EXPERIMENTAL RESULTS

In this study, taking the multiple STCT geometry of the LCT system as a typical example, some preliminary configurations are set as follows: The unit size of the detector is 0.127 mm; $l$ *and* $h$ are 15 mm and 190 mm, respectively; $T$ is five segments; $\Delta\theta$ is 36.5°; the corresponding calculations can be found in Refs. [5], [7], [14], [50].

We compared OSNet and MNetO and marked their peak signal-to-noise ratio (PSNR) and structural similarity (SSIM) in Fig. 5. Compared to BPF, both OSNet and MNetO have higher clarity, and OSNet is closer to the original image with lower root mean square error (RMSE). Moreover, the average PSNR of OSNet (41.1949 dB) shown in Table I is much higher than that of MNetO (22.3197 dB).

MNeto can restore the final complete image, but it can easily generate some artifacts and pixel migration, as shown in Fig. 5. Though MNetO is inferior to OSNet, but it exhibits good generalization performance in a single model. Specifically, we test the DBP images of other types of CT images (i.e., OOD slices) outside the training set, as shown in Fig. 6. MNetO performs relatively poorly compared to OSNet in the final reconstructed image effect, but RMSE results (ST-pix2pixGAN:0.0096) on each segment using multiple model learning for a single-direction Hilbert inverse process are better than the RMSE results (ST-pix2pixGAN: 0.0134) of OSNet on the superimposed DBP image. An important application of MNetO is that the area we need to detect information is close to the interior of an edge, so we only need one-segment STCT with a special direction rather than more doses of OSNet. In Fig. 7, we simulate the contour information and learn the Hilbert filtering of the first segment of the DBP image using one of the models in MNetO. Compared with BPF, MNetO based on ST-pix2pixGAN can present this type of detail more clearly.

Generally, a complete reconstruction is required to investigate more details within a large FOV. Therefore, in the next methods and experiments, to better demonstrate the advantages and characteristics of our method, we selected OSNet for experiments with different networks (including pix2pixGAN, CycleGAN, and ST-pix2pixGAN.

Fig. 8 shows the results and corresponding profile curves using OSNet under different numbers of reconstructed pixels. The scenes with 256*256, 512*512, and 1024*1024 reconstructed pixels were selected for comparison of the scan results of the same object with the same scan, and the corresponding pixel sizes are 33.1 μm, 16.5 μm, and 8.3 μm. When the reconstructed image size is smaller, the image obtained by the BPF algorithm is blurrier, while the deep learning methods perform better under the same conditions. According to the profile curves, the fewer the number of reconstructed pixels, the farther the result obtained by BPF deviates from the original curve, while ST-pix2pixGAN can consistently achieve the best image quality in any case. Table II lists the average PSNR, SSIM, and RMSE corresponding to different numbers of reconstructed pixels.

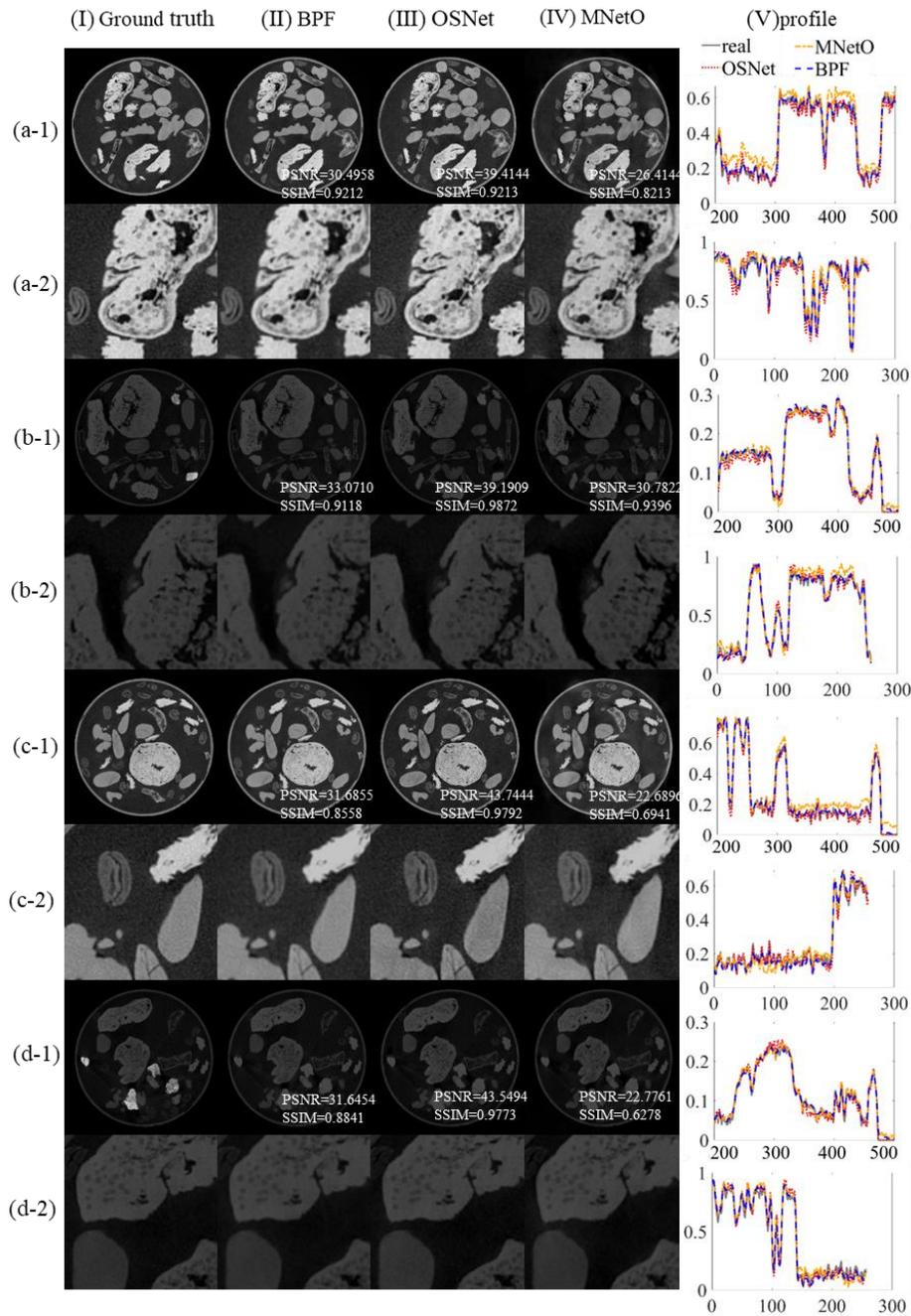

**Fig. 5.** Column direction: Ground-truth(I), BPF algorithm (II), ST-pix2pixGAN was used to obtain OSNet (III) and MNetO (IV) images (a/b/c/d-1)(1024*1024) with their central horizontal profiles (V) and locally enlarged images (a/b/c/d-2)(256 * 256) with their central vertical profiles, respectively.

TABLE I

COMPARISON CHART OF EVALUATION INDICATORS FOR METHODS OSNET AND MNETO

| metric | BPF | OSNet | MNetO |
|---|---|---|---|
| PSNR (dB) | 31.4415 | 41.1949 | 22.3197 |
| RMSE | 0.0225 | 0.0093 | 161.00 |
| SSIM | 0.9987 | 0.9996 | 0.0805 |

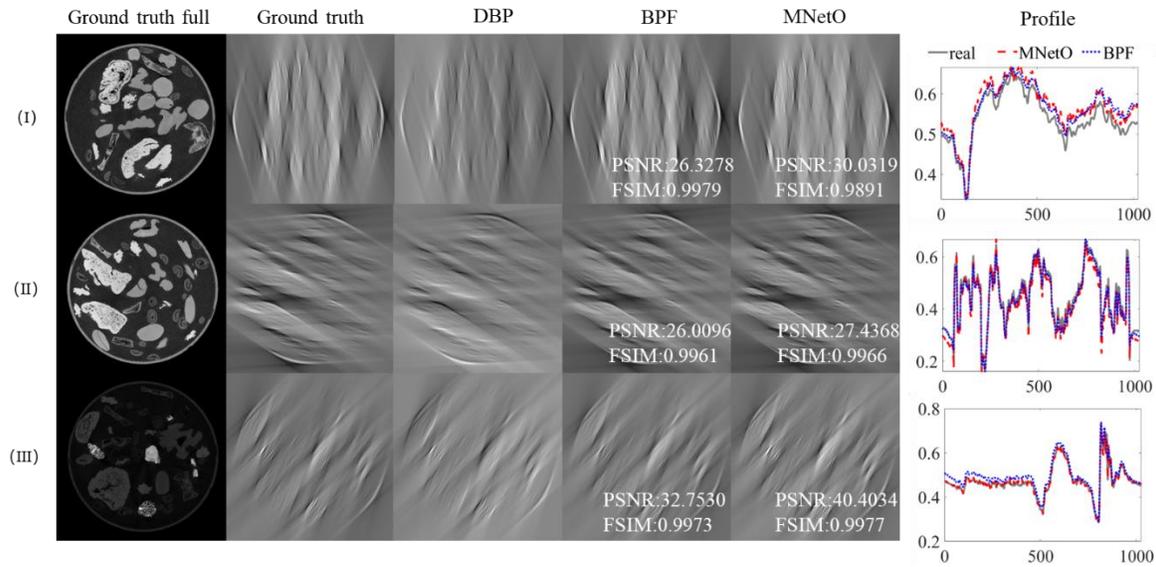

**Fig. 6.** Performance of different images in MNetO: (I):1st STCT; (ii): 2nd STCT; (III): 4th STCT. On the far right is the central horizontal profile curve for each row of contrast images.

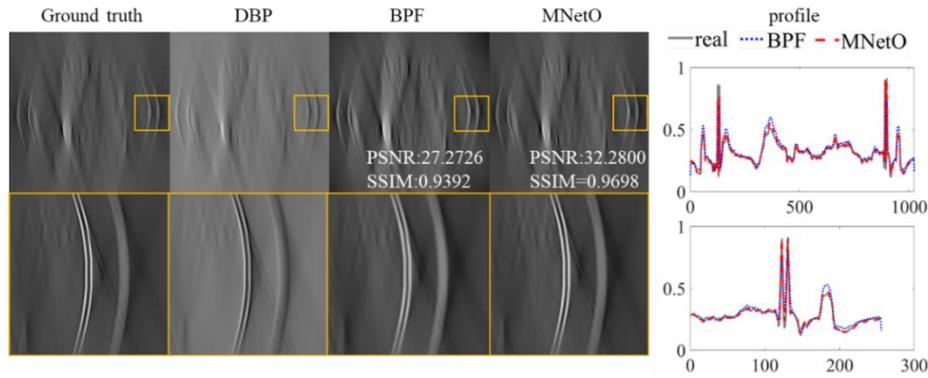

**Fig. 7.** Comparison of individual image details in MNetO. Note that MNetO can present a sharper boundary than BPF from figures. The rightmost column gives the central horizontal profiles of each row of images.

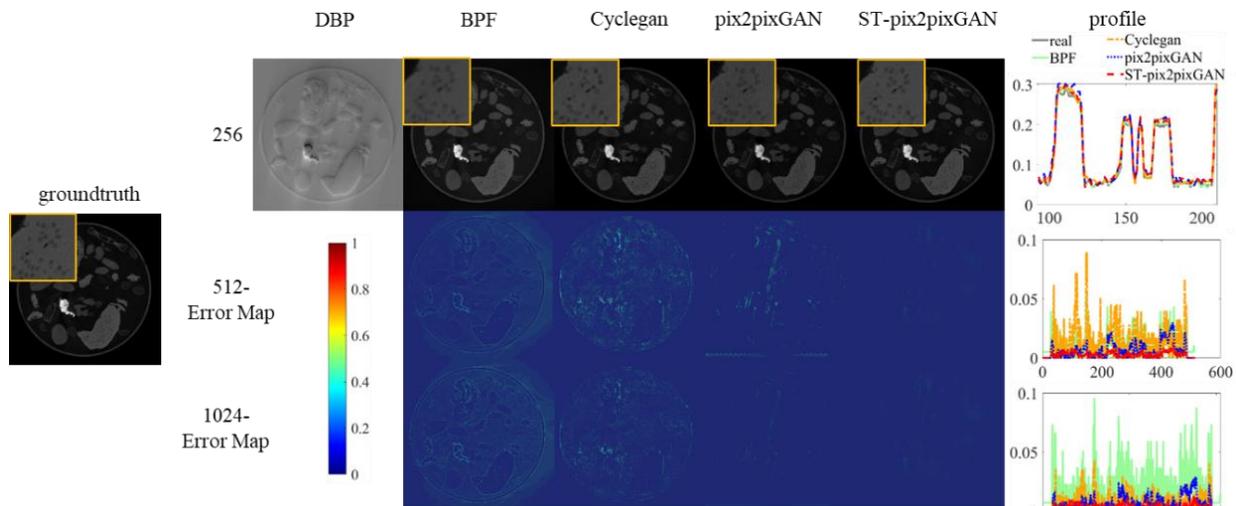

**Fig. 8.** BPF, CycleGAN, pix2pixGAN, and ST-pix2pixGAN reconstruct images, residuals, and profile curves in OSNet. The first row displays reconstruction images with the number of pixels 256 * 256 and their central horizontal profiles from the 100th pixel to the 200th pixel. The second and third rows display reconstruction error images with the numbers of pixels 512 * 512 and 1024 * 1024, respectively, and their central horizontal profiles.

Fig. 9 presents the experimental results between OSNet and BPF under different lengths of linear trajectory *LS*, from 12 mm to 20 mm (such as 12 mm, 14 mm, 16 mm, 18 mm, and 20 mm, setting object size equal to the FOV size with the length of linear trajectory being 20 mm, i.e., changing *LS* can determine different FOV sizes, and the detailed calculation can be found in Refs. [7], [14]). We produce separate data sets for different FOV regions. When the length of the linear trajectory is 18 mm or smaller, BPF will yield unstable reconstructed results and artifacts, leading to a decrease in image quality, while OSNet can avoid this problem. When the length of the linear trajectory is 12 mm, the deep learning method hardly produces artifacts and can show clearer results. The PSNR of ST-pix2pixGAN can reach 37.7365 dB, while BPF is only 13.6598 dB. OSNet shows good robustness in various linear scanning scenes. Fig. 10 shows the evaluation indicators of experimental results under different lengths of linear trajectory.

Fig. 11 displays the comparison between OSNet and BPF under different numbers of source sampling points per linear trajectory, including 251, 1001, and 2001. There is little difference between the algorithms. All types of deep learning networks under OSNet can reconstruct the original image well. Table III lists the performance of various deep learning methods under BPF and OSNet. CycleGAN performs unstably, but the pix2pixGAN method can reconstruct high-quality images, and the reconstruction quality gradually improves with the increase of source sampling points.

TABLE II

COMPARISON OF VARIOUS EVALUATION INDICATORS WHEN THE NUMBER OF PIXELS TO BE RECONSTRUCTED IS 1024

|  | | OSNet | | |
| --- | --- | --- | --- | --- |
|  | BPF | Cyclegan | pix2pixGAN | ST-Pix2pix |
| **PSNR** | | | | |
| 256 | 24.4328 | 34.0972 | 37.0481 | 35.8425 |
| 512 | 27.9174 | 35.8063 | 37.4855 | 39.9043 |
| 1024 | 31.9568 | 36.8697 | 40.7205 | 41.6542 |
| **SSIM** | | | | |
| 256 | 0.6952 | 0.9417 | 0.9753 | 0.9465 |
| 512 | 0.7520 | 0.9403 | 0.9634 | 0.9744 |
| 1024 | 0.8264 | 0.9360 | 0.9747 | 0.9802 |
| **RMSE** | | | | |
| 256 | 0.0302 | 0.0172 | 0.0156 | 0.0123 |
| 512 | 0.0148 | 0.0157 | 0.0137 | 0.0101 |
| 1024 | 0.0123 | 0.0172 | 0.0126 | 0.0096 |

TABLE III

COMPARISON OF AVERAGE EVALUATION INDICATORS FOR 1001 SOURCE MOVEMENT STEPS

|  | | OSNet | | |
| --- | --- | --- | --- | --- |
|  | BPF | Cyclegan | pix2pixGAN | ST-pix2pixGAN |
| **PSNR** | | | | |
| 128 | 31.9614 | 36.7307 | 39.8532 | 39.7903 |
| 256 | 31.9596 | 33.8379 | 39.4457 | 40.6720 |
| 512 | 31.9568 | 40.3415 | 39.4201 | 42.5188 |
| **SSIM** | | | | |
| 128 | 0.8261 | 0.9342 | 0.9653 | 0.9568 |
| 256 | 0.8271 | 0.8336 | 0.9624 | 0.9636 |
| 512 | 0.8270 | 0.9605 | 0.9630 | 0.9837 |
| **RMSE** | | | | |
| 128 | 42.4893 | 25.8 | 21.2 | 24 |
| 256 | 42.4924 | 51.6 | 21.8 | 21.8 |
| 512 | 42.2976 | 17.2 | 22 | 16.4 |

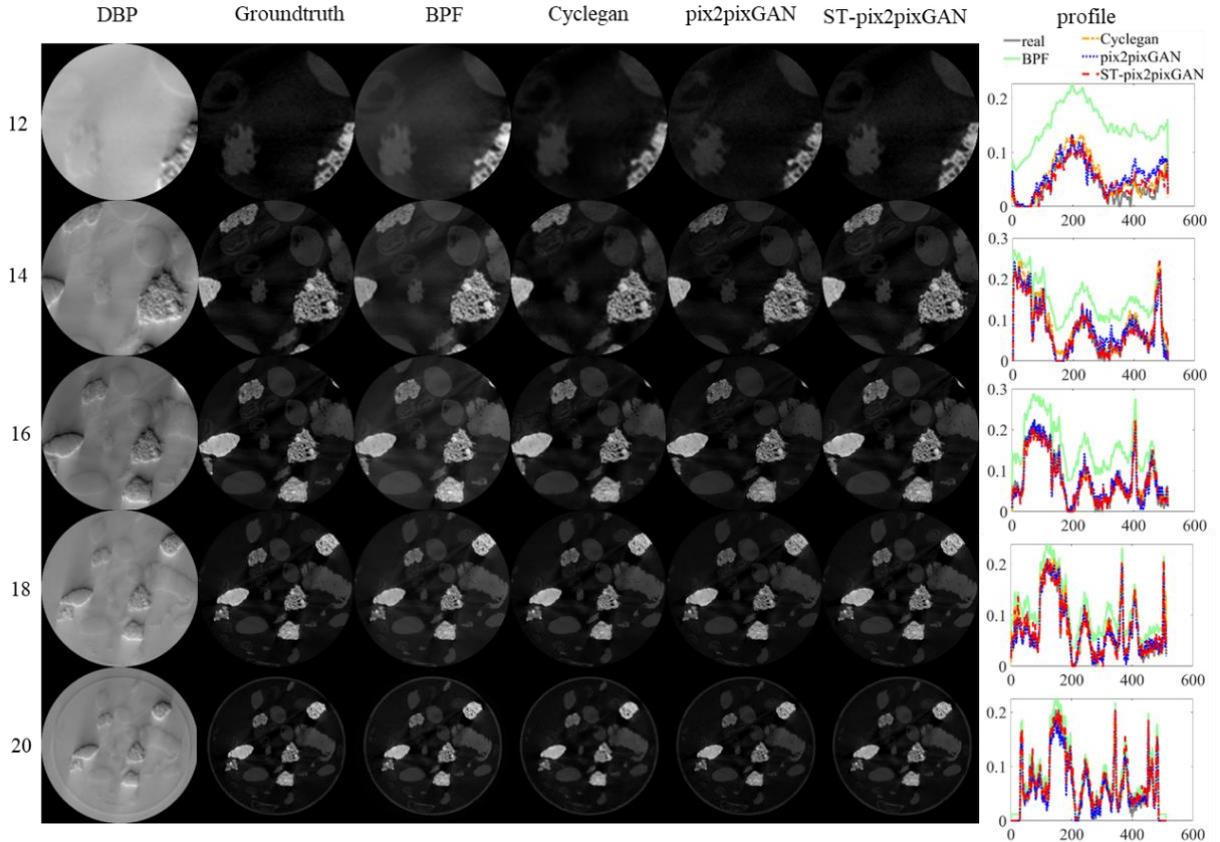

**Fig. 9.** Reconstructions among CycleGAN, pix2pixGAN, ST-pix2pixGAN, and BPF in OSNet with the length of the linear trajectory *LS* from 12 mm to 20 mm, including 12 mm, 14 mm, 16 mm, 18 mm, and 20 mm, respectively. The rightmost column describes the central horizontal profiles of each row of images.

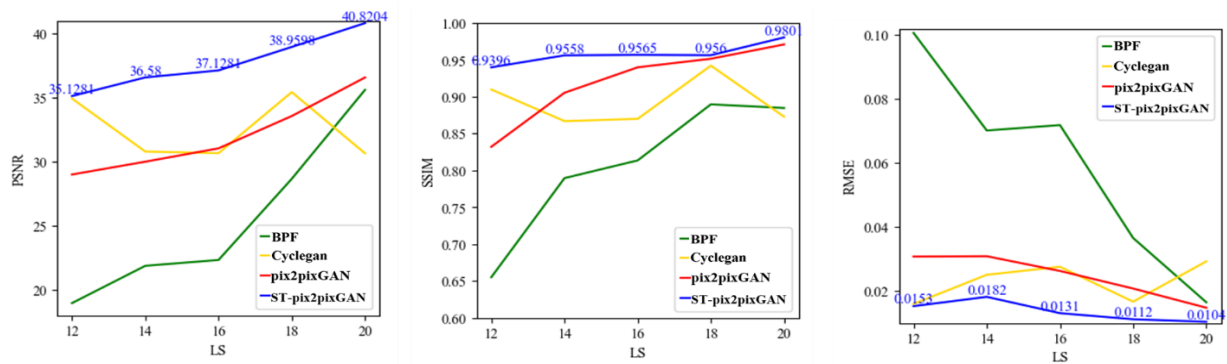

**Fig. 10.** Comparison of PSNR, SSIM, and RMSE of various methods with the length of the linear trajectory *LS* from 12 mm to 20 mm.

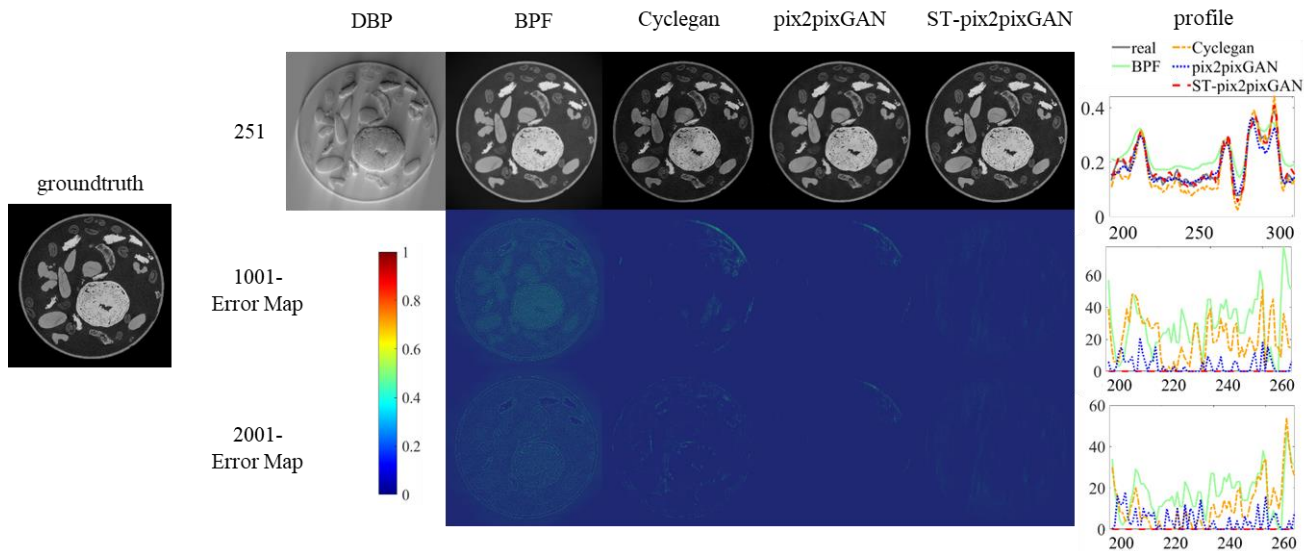

**Fig. 11.** BPF, CycleGAN, pix2pixGAN, and ST-pix2pixGAN reconstruct images, residuals, and profiles in OSNet. The first row shows the reconstruction images and their central horizontal profiles from 200th to 300th pixels with 251 STCT source sampling points. The second and third rows exhibit error maps between each reconstruction image and the ground truth, with source sampling points of 1001 and 2001 and partial central horizontal profiles, respectively.

The geometric magnification ratio refers to the ratio of the distance between the source and the detector plane to the distance between the source and the rotation center. While adjusting the amplification ratio, to maintain the FOV size unchanged, we set the size of the measured object to be equal to the diameter of the FOV. Fig. 12 describes the evaluation indicators at different geometric magnification ratios. At low geometric magnification ratios, such as 110 and 120, the images reconstructed by the BPF algorithm show lower quality, while OSNet performs better.

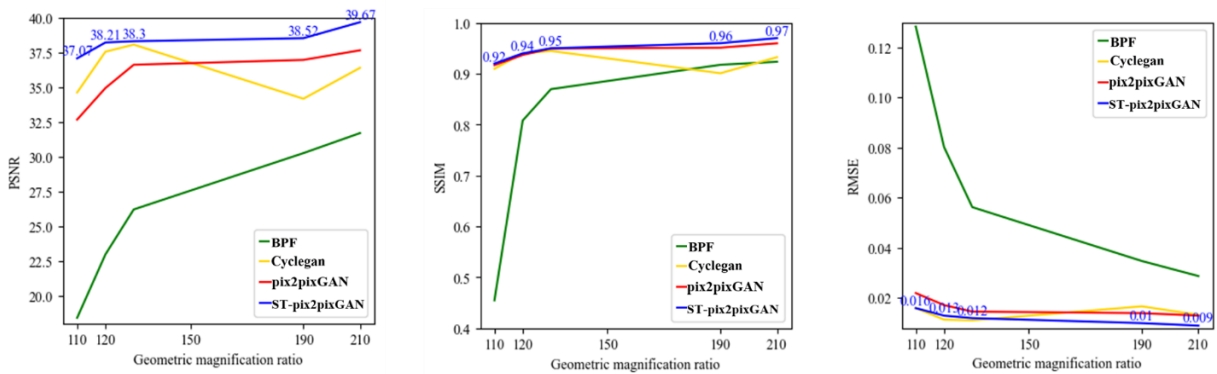

**Fig. 12.** Comparison of PSNR, SSIM and RMSE of different methods in the range of 110 mm - 210 mm from the detected sample.

In addition, we randomly selected some slices from the 750 OOD slices in the **2DeteCT** dataset as inputs to verify the generalizability of our proposed OSNet. The test results are shown in Fig. 13. Note that OSNet was still able to obtain high-quality reconstructed images. The profiles on the rightmost of Fig. 13 show that BPF lose some of the sharp edge details, resulting in loss of high-frequency information in the image, and OSNet can reconstruct these details well.

Finally, Table IV lists the computation time of various methods at different reconstruction image sizes. The proposed method only requires 0.068 s/slice for OSNet (single model) to run at a resolution of 128 pixels, and can be accelerated on GPU. The runtime of MNetO is a multiple of the single model and is directly related to the number of scanning sections. All kinds of deep learning can achieve higher efficiency than BPF algorithm when using trained models for image reconstruction, among which our method ST-pix2pixGAN can achieve faster reconstruction speed than traditional pix2pixGAN when processing large-size images (512, 1024).

TABLE IV
COMPARISON OF RECONSTRUCTION TIME FOR MULTIPLE METHODS

| Time s/slice | OSNet | | | | MNetO | |
|---|---|---|---|---|---|---|
| | BPF | Cycle GAN | pix2pix GAN | ST-pix2pix GAN | pix2pix GAN | ST-pix2pixG AN |
| 128 | 1.70 | 0.08 | 0.06 | 0.08 | 0.32 | 0.37 |
| 256 | 1.73 | 0.23 | 0.12 | 0.17 | 0.54 | 0.63 |
| 512 | 1.89 | 0.56 | 0.25 | 0.22 | 1.21 | 1.19 |
| 1024 | 2.45 | 1.69 | 0.28 | 0.27 | 1.69 | 1.56 |

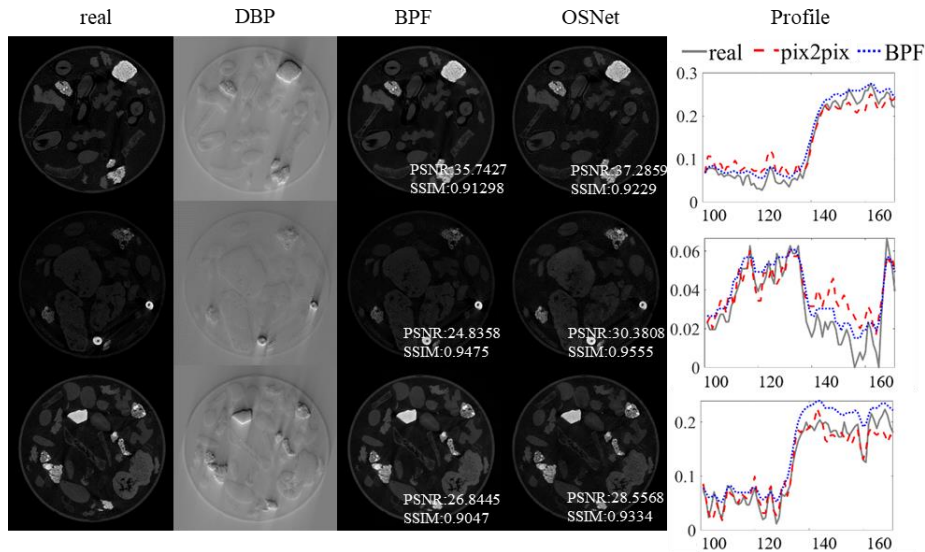

**Fig. 13.** The experimental results of parts of OOD slices selected to test the generalizability of OSNet. The rightmost column shows the central horizontal profiles of each row of images from 100 to 160 pixels.

## V. DISCUSSION

### A. Discussion on OSNet

Note that in our first approach, OSNet, whether to use a paired form has a crucial impact on the reconstruction results of conditional adversarial networks. For example, in different scenarios of the length of linear trajectory (i.e., FOV size) in Fig. 9 or different geometric magnification ratios in Fig. 12, both pix2pixGAN and ST-pix2pixGAN methods perform well, while CycleGAN presents unstable characteristics. Recall that our method uses a paired approach for image generation. If the image domain style transformation is used during training instead of one-to-one image transformation, the model will inevitably learn the features of specificities of other images.

From the tested results of some OOD slices from the **2DeteCT** dataset shown in Fig. 13, we can notice that OSNet still displays great generalizability. This is of great significance for the portability of deep learning models and the engineering of this method. Our findings suggest that the OSNet architecture we propose has the ability to learn a dimensionless functional mapping between inputs (superimposed and complex DBP images of multiple linear scans in LCT) and outputs (final object functions), rather than relying on specific weighting functions with precise physical units. Therefore, it could potentially explain the generalization ability of OSNet in our experiments for test images with varying field-of-view sizes, pixel sizes, number of projections, and geometric magnification.

### B. Discussion on MNetO

MNetO conforms to the algorithm process of BPF for LCT, but it cannot obtain better results because some slight differences exist between different trained models that cannot be mutually offset most of the time, resulting in artifacts in the final reconstructed images. As shown in Fig. 6, the profiles also show that the red curve is closer to the original. The single DBP image of various types of images can perform well in MNetO, and it is sharper compared to one of BPF. This implies that our networks learned a cluster of parallel-lines Hilbert filtering with different directions in theory, and the superimposed DBP image reconstruction is essentially a model mapped by overlapped Hilbert filtering functions with different directions. In MNetO, the deep learning model can more easily fit this type of single-directional solution space to achieve image transformation. Therefore, when replacing different types of images, the model can still exhibit strong generalization.

Any one of the models of MNetO cannot obtain the complete image, but it is suitable for high-quality exterior reconstruction in a certain direction, especially in inspecting the inner wall of the in-service pipelines.

### C. Discussion on ST-pix2pixGAN

Our proposed ST-pix2pixGAN utilizes the Transformer's attention mechanism as the encoder in image generation, replacing the convolutional layers in the traditional U-Net. This design brings several advantages. Firstly, the attention mechanism better captures global information and reduces information loss in feature maps. Secondly, the attention mechanism is effective in handling features at different scales. In DBP images of LCT, there are often significant global pixel shifts, particularly in areas with higher pixel values. Each repeated layer in Swin Transformer can fully extract both local and global features of the images. From the residual images, it can be observed that when the traditional U-Net is used as the generator in the image generation model, it struggles to learn the global feature space well. This leads to significant pixel shifts in generated images and residual artifacts in the residual images. In contrast, the improved ST-pix2pixGAN network can learn features at different scales more effectively, almost eliminating pixel shift issues.

In the OSNet architecture, the ST-pix2pixGAN network is more suitable for scenarios with large image sizes or a larger number of detector pixels. Because ST-pix2pixGAN uses multi-scale generation and discrimination as well as image semantic editing technology, it can generate more detailed image details, but at the cost of a longer training time (At 512*512 pixels, ST-pix2pixGAN takes 11.2 hours per model, while pix2pixGAN takes 4.9 hours per model). In the future, we will also try to streamline the model or use techniques such as pruning or parameter sharing to improve the efficiency of model training.

### D. Analysis and comparsion between OSNet and MNetO

In fact, the purpose of OSNet is to let the deep learning model learn the complex texture space of multiple overlapping DBP images about finite Hilbert filtering. For more complex medical images, the model inevitably learns some prior information, leading to local overfitting and poor performance when testing images with different structural types. To some extent, the multiple parallel finite Hilbert filter function for different directions is more complex and prone to overfitting of the learning model, while the single-directional parallel finite Hilbert filter function is relatively simple, and the deep learning model is easier to learn and less prone to overfitting in MNetO, so its generalization ability should be better.

### V. CONCLUSION

In this work, we propose two types of general reconstruction architectures, i.e., OSNet and MNetO, for LCT reconstruction in multi-scenarios, including interior ROI, complete object, and exterior edge, avoiding the rotation-Hilbert filtering-inverse rotation operation for each DBP image. Our proposed network architectures can greatly learn the Hilbert filtering function parallel to different linear trajectories to translate the DBP images of LCT to the reconstructed objects. OSNet can achieve high-quality, complete, and even interior reconstruction. OSNet shows better performance compared to MNetO and can be extended to more scenarios. Among various networks, ST-pix2pixGAN exhibited the best performance. Massive experiments verify that OSNet outperformed BPF in terms of image quality as well as computation time. MNetO shows some artifacts due to the differences in the multiple models, but any one of its models is suitable for high-quality exterior reconstruction in a certain direction.

### ACKNOWLEDGMENT

This research is supported by National Natural Science Foundation of China (Grant No.: 52075133), CGN-HIT Advanced Nuclear and New Energy Research Institute (Grant No.: CGN-HIT202215).